\documentclass[a4paper, 11pt]{report}
\usepackage[utf8]{inputenc}
\usepackage{amssymb, amsmath}
\usepackage{graphicx} % figure
\usepackage{booktabs} % table
\usepackage{algorithm} % pseudo algorithms
\usepackage{algpseudocode} % pseudo algorithms
\usepackage[nottoc]{tocbibind} % add bibliography to the table of contents
\graphicspath{{images/}}
\title{\textbf{Convolutional Attention-based Seq2Seq Neural Network for End-to-End ASR}}
\author{\textbf{Dan Lim} \\ Thesis for the Degree of Master \\ 
Department of Computer Science and Engineering \\
Graduate School Korea University}
\date{}

\begin{document}

% titlepage adapted from https://en.wikibooks.org/wiki/LaTeX/Title_Creation
\begin{titlepage}
	\centering
	{\scshape\LARGE Thesis for the Degree of Master \par}
	\vspace{1cm}
	%{\scshape\Large Final year project\par}
	%\vspace{1.5cm}
	{\huge\bfseries Convolutional Attention-based Seq2Seq Neural Network for End-to-End ASR\par}
	\vspace{2cm}
	{\Large\itshape by \par Dan Lim}
	\vfill
	{Department of Computer Science and Engineering\par
	Korea University Graduate School}

	\vfill

% Bottom of the page
	%{\large \today\par}
\end{titlepage}

\begin{abstract}
\setcounter{page}{2} % abstract reset page number at the start, end of environment
\thispagestyle{plain} % empty(page number not shown) -> plain

Traditional approach in artificial intelligence (AI) have been solving the problem that is difficult for human but relatively easy for computer if it could be formulated as mathematical rules or formal languages. However, their symbol, rule-based approach failed in the problem where human being solves intuitively like image recognition, natural language understanding and speech recognition.

Therefore the machine learning, which is subfield of AI, have tackled this intuitive problems by making the computer learn from data automatically instead of human efforts of extracting complicated rules. Especially the deep learning which is a particular kind of machine learning as well as central theme of this thesis, have shown great popularity and usefulness recently.

It has been known that the powerful computer, large dataset and algorithmic improvement have made recent success of the deep learning. And this factors have enabled recent research to train deeper network achieving significant performance improvement. Those current research trends motivated me to quest deeper architecture for the end-to-end speech recognition. 

In this thesis, I experimentally showed that the proposed deep neural network achieves state-of-the-art results on `TIMIT' speech recognition benchmark dataset. Specifically, the convolutional attention-based sequence-to-sequence model which has the deep stacked convolutional layers in the attention-based seq2seq framework achieved 15.8\% phoneme error rate.

\end{abstract}

\setcounter{page}{3}

\tableofcontents

\clearpage
\pagenumbering{arabic}

\chapter{Introduction}

The deep learning is the subfield of machine learning where computer learns from data while minimizing the human intervention. It is loosely inspired by the human brain and quiet special in that it learns abstract concepts with hierarchy of concepts by building them out of simpler ones. It is called deep learning since if one draw graph describing model, the graph becomes deep, with many stacked layers where upper layers exploit representation of lower layers to make more abstract concepts.

The deep learning approach are quiet successful in various domain including image recognition, machine translation and speech recognition. For example, deep residual network \cite{he2016deep} consists of more than a hundred convolutional layers winning ILSVRC 2015 classification task, Google \cite{DBLP:journals/corr/WuSCLNMKCGMKSJL16} have been providing neural machine translation service and Deep Speech 2 \cite{pmlr-v48-amodei16} is successfully trained deep learning based end-to-end speech recognition system in English and Chinese speech.

In the case of the end-to-end automatic speech recognition, the attention-based sequence to sequence model (attention-based seq2seq) are the active research area currently. It already have shown the flexibility as general sequence transducer in various domain including machine translation \cite{Bahdanau2016Neural}, image captioning \cite{Xu2016Show}, scene text recognition \cite{DBLP:journals/corr/WojnaGL0YLI17}, as well as speech recognition \cite{chan2016listen}. 

Nevertheless I suspected there are potential benefit if it has deeper networks. For example, end-to-end ASR based on deep convolutional neural network \cite{Zhang2016Very}, \cite{DBLP:journals/corr/ZhangPBZLBC17}, \cite{wang2017residual}, have shown the superiority of deep architecture as well as effectiveness of convolutional neural network for speech recognition. This prior works are similar with the proposed model in this thesis in that they used deep convolutional neural networks.

However, there were other various algorithmic configurations that should be considered in the proposed model. This includes training deep convolutional neural network without performance degradation with Residual network \cite{he2016deep}, Batch normalization \cite{ioffe2015batch}, regularizing deep neural network with Dropout \cite{Srivastava:2014:DSW:2627435.2670313}, deciding effective attention mechanism of seq2seq model with Luong's attention method \cite{DBLP:journals/corr/LuongPM15}.

So how to combine this algorithms to build deep neural network for speech recognition has been the remained questions for deep learning researchers. In this thesis, I proposed the convolutional attention-based seq2seq neural network for end-to-end automatic speech recognition. The proposed model utilizes structured property of speech data with deep convolutional neural network in the attention-based seq2seq framework. In the experiment on `TIMIT' speech recognition benchmark dataset, the convolutional attention-based seq2seq model achieves state-of-the-art results; 15.8\% phoneme error rate.

\section{Thesis structure}

This thesis is organized as follows:

\begin{itemize}
\item Chapter 2 explain about automatic speech recognition, the acoustic features and introduce the neural network. Then it explain attention-based sequence to sequence model which is the main framework for end-to-end ASR.

\item Chapter 3 describe each components of the proposed model including Luong's attention mechanism, Batch normalization, Dropout and Residual network. These are combined to complete the proposed model.

\item Chapter 4 explain the proposed model architecture, dataset, training and evaluation method in great details. Then the experiment result is provided as compared to the prior research results on same dataset.

\item Chapter 5 make conclusions of my thesis and discuss future research directions.
\end{itemize}

\section{Thesis contributions}

This thesis introduces the sequence to sequence model with Luong's attention mechanism for end-to-end ASR. It also describes various neural network algorithms including Batch normalization, Dropout and Residual network which constitute the convolutional attention-based seq2seq neural network. Finally the proposed model proved its effectiveness for speech recognition achieving 15.8\% phoneme error rate on `TIMIT' dataset which is state-of-the-art result as far as I know.

\chapter{Background}

\section{Automatic speech recognition}

Automatic speech recognition (ASR) is the problem of identifying intended utterance from human speech. Formally, given speech in the form of acoustic feature sequences $\mathbf{x} = (x_1, x_2, ..., x_T)$, most probable word or character sequences $\mathbf{y} = (y_1, y_2, ..., y_{T'})$ would be found as:
\begin{equation}
\hat{\mathbf{y}} = \arg \max_{\mathbf{y}} P(\mathbf{y}|\mathbf{x})
\end{equation}
Where $P(\mathbf{y}|\mathbf{x})$ is the conditional distribution relating input $\mathbf{x}$ to the outputs $\mathbf{y}$.

Until about 2009-2012, state-of-the-art speech recognition system used GMM-HMM (Gaussian Mixture Model - Hidden Markov Model). The HMM modeled the sequences of state which denotes phoneme (basic unit of sound) and GMM associated the acoustic feature with the state of HMM. Although the ASR based on neural network was suggested and had shown comparable performance of GMM-HMM systems, the complex engineering involved in software systems on the basis of GMM-HMM made it standard in the industry for a long time.

Later with much larger models and powerful computers, the ASR performance was dramatically improved by using neural networks to replace GMM. For example, \cite{5704567} showed that DNN-HMM (Deep Neural Network - Hidden Markov Model) improves the recognition rate of `TIMIT' significantly, bringing down the phoneme error rate from about 26\% to 20.7\%. Where `TIMIT' is a benchmark dataset for phoneme recognition, playing a similar role of `MNIST' used for handwritten digits recognition.

Today active research area is the end-to-end automatic speech recognition based on deep learning that completely remove HMM with single large neural network. For example, the connectionist temporal classification (CTC) \cite{graves2006connectionist} allows networks to output blank and repeated symbols and map target sequences which has length not greater than input sequences by marginalize all possible configuration.

Another end-to-end ASR system is attention-based sequence to sequence model which learns how to align input sequences with output sequences. It already have achieved comparable results to prior research on phoneme recognition \cite{chorowski2015attention}, giving 17.6\%. The attention-based sequence to sequence model is special interest of this thesis and one of the key components of the propose model; Convolutional attention-based seq2seq neural network.

\section{Acoustic features}

The first step for training the speech recognition model is to extract acoustic feature vector from speech data. In this thesis, I used log mel-scaled filter banks as speech feature vector. It has been known to preserve the local correlations of spectrogram so it is mainly used in the end-to-end speech recognition research.

The feature extraction step could be described in the following orders:

\begin{enumerate}
\item Although the speech signal is constantly changing, it could be assumed to be statistically stationary in short time scale. So the speech data is divided into 25ms overlapping frame at every 10ms.

\item Compute power spectrum of each frame by applying Short Time Fourier Transform (STFT). This is motivated by human cochlea which vibrates at different spot depending on the frequency of incoming sound.

\item To reflect the perception mechanism of human sound that is being more discriminative at lower frequencies and less discriminative at higher frequencies, the mel-scaled filter bank is applied to the power spectrum of each frame. Where the filter banks indicate how much energy exists in each frequency band and the mel scale decides how to space out filter banks and how wide to make them.

\item Finally the log mel-scaled filter bank coefficients are obtained by taking logarithm of mel-scaled filter bank energies. This is also motivated by human hearing that sound loudness is not perceived linearly.
\end{enumerate}

Since the log mel-scaled filter bank coefficients only describe power spectral envelope of single frame, the delta and delta-delta coefficients which are also known as differential and acceleration coefficients are often appended to log mel-scaled filter bank features.

The delta coefficient at time $t$ is computed as:

\begin{equation}
d_t = \frac{\sum_{n=1}^N n(c_{t+n} - c_{t-n})}{2 \sum_{n=1}^N n^2}
\end{equation}

Where $d$ is delta coefficient, $c$ is log mel-scaled filter bank coefficient and typical value of $N$ is 2. The delta-delta coefficient is calculated in the same way starting from the delta features.

Although not used in this thesis, the mel frequency cepstral coefficient (MFCC) is also frequently used feature vector for ASR. It is obtained by applying discrete cosine transform (DCT) to log mel-scaled filter bank and keeping the 2-13 coefficients. The motivation of it is to decorrelate the filter bank coefficients since the overlapping filter bank energies are highly correlated. 

\section{Neural network}

Neural network is specific application of deep learning. To understand the proposed model which is end-to-end speech recognition system based on neural network, the minimum backgrounds are provided including the type of neural network, optimization and regularization.

\subsection{Feed forward neural network}

\begin{figure}[h!]
\centering
\includegraphics[width=0.5\textwidth,height=0.5\textwidth]{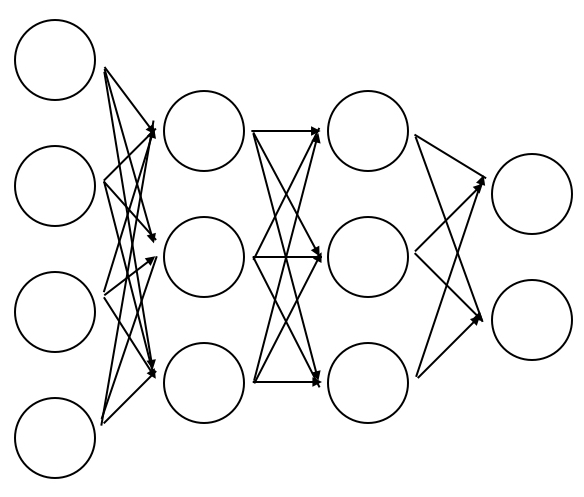}
\caption{The feed forward neural network: example of one input layer, two hidden layers and output layer of two nodes}
\label{fig:feedforward}
\end{figure}

Most basic form of neural network is feed forward neural network where input signal flows in forward direction without feedback connection. It is also called fully connected or dense neural network in that each output value is computed from all input values. 

The operation performed in feed forward neural network is affine transform followed by non linear activation function. Given vector valued input $\mathbf{x}$, it compute new representation $\mathbf{h}$ as:

\begin{equation}
\mathbf{h} = \mathcal{F}(W \mathbf{x} + b)
\end{equation}

Where $\mathcal{F}$ is element-wise nonlinear activation function, $W$ is weight matrix and $b$ is bias parameter. 

It is expected to be able to obtain more suitable representations for final output by composing above function multiple times. The composed functions could be depicted as shown \ref{fig:feedforward}. In this sense, it is called the stacked neural layers or deep learning.

The activation function made neural network become a powerful function approximator by giving it a non linearity. The one of the commonly used activation function is Rectified Linear Unit (ReLU) which is easy to optimize since it has similar property to linear unit.

\begin{equation}
y = ReLU(x) = \max (0, x)
\end{equation}

\subsection{Convolutional neural network}

One of main contributions of this thesis is to prove effectiveness of convolutional neural network (CNN) in the attention-based seq2seq model. The CNN have been known to be good at handling structured data like image with weight sharing and sparse connectivity. Recently successfully trained very deep CNN have shown human level performance on image recognition dataset.

The acoustic feature vector in the form of log mel-scaled filter bank preserves local correlations of the spectrogram so CNN is expected to be good at modeling this property rather than feed forward neural network.

Specifically in the case of speech dataset, given acoustic feature vector sequences $\mathbf{X} \in \mathbb{R}^{f \times t \times c}$ with frequency band width $f$, time length $t$ and channel depth $c$, the convolutional neural network convolves $\mathbf{X}$ with $k$ filters $\{\mathbf{W}_i\}_k$ where each $\mathbf{W}_i \in \mathbb{R}^{m \times n \times c}$ is a 3D tensor with frequency axis equals $m$, time axis equals $n$. The resulting $k$ preactivation feature maps consist of a 3D tensor $\mathbf{H} \in \mathbb{R}^{f_H \times t_H \times k}$ in which each feature map $\mathbf{H}_i$ is computed as follows:

\begin{equation}
\mathbf{H}_i = \mathbf{W}_i * \mathbf{X} + b_i, \quad i = 1, ..., k.
\end{equation}

The symbol $*$ is convolution operation and $b_i$ is a bias parameter.

\subsection{Recurrent neural network}

\begin{figure}[h!]
\centering
\includegraphics[width=\textwidth,height=0.5\textwidth]{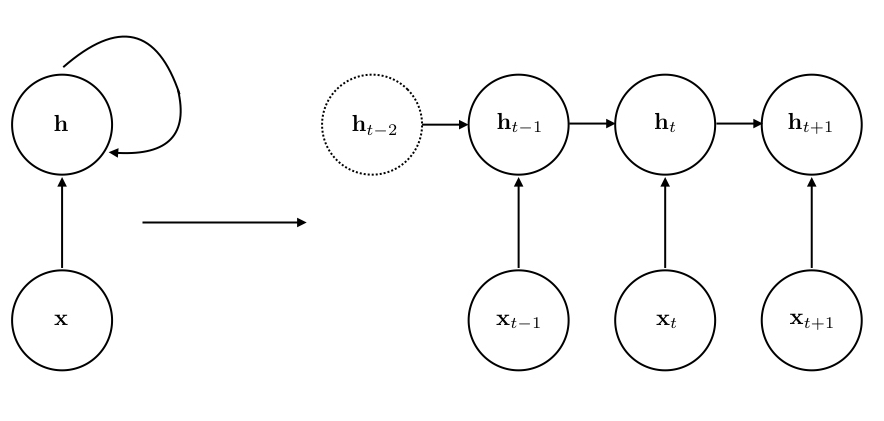}
\caption{(Left) recurrent neural network. (Right) recurrent neural network as an unfolded computational graph.}
\label{fig:rnn}
\end{figure}

Recurrent neural network (RNN) has feedback connection as shown in figure \ref{fig:rnn} and this feedback connection makes RNN to process variable-length sequences like acoustic feature vector sequences in speech recognition. 

The computed value at each time step in RNN is hidden state. For example, given hidden state $h_{t-1}$ and $x_t$ input values, it compute next hidden state as:

\begin{equation}
h_{t} = \mathcal{F}(W_h h_{t-1} + W_x x_{t} + b)
\end{equation}

Where $\mathcal{F}$ is nonlinear activation function like $tanh$.

This way, the RNN processes variable-length sequences one at a time and if RNN is used for predicting next value from past sequences, hidden state $h_t$ acts as lossy summary of past input sequences $x_{<t}$.

However in practice, standard RNN is hard to successfully train if input sequences have long length which is common in speech recognition. Like deep feed forward neural network is difficult to train, the RNN with long sequences becomes deep stacked layer in time axis that causes the signal from the past inputs vanished easily. In perspective of neural network optimization, it is called gradient vanishing problem.

So instead of standard RNN, the proposed model used Long Short Term Memory (LSTM) which is widely used in speech recognition domain \cite{Graves2013Speech}. The LSTM uses gating mechanism to cope with gradient vanishing problem. By having internal recurrence where memory cell state $c_t$ could propagate to next time step in linear fashion rather than affine transformation, it could learn long term dependencies more easily. Specifically, LSTM with peephole connection is what I used in this thesis and it computes hidden state as follows:

\begin{equation}
\begin{aligned}
i_t &= \sigma(W_{xi} x_t + W_{hi} h_{t-1} + W_{ci} c_{t-1} + b_i) \\
f_t &= \sigma(W_{xf} x_t + W_{hf} h_{t-1} + W_{cf} c_{t-1} + b_f) \\
c_t &= f_t c_{t-1} + i_t tanh(W_{xc} x_t + W_{hc} h_{t-1} + b_c) \\
o_t &= \sigma(W_{xo} x_t + W_{ho} h_{t-1} + W_{co} c_t + b_o) \\
h_t &= o_t tanh(c_t)
\end{aligned}
\end{equation}
Where $\sigma$ is logistic sigmoid function, and $i_t, f_t, o_t, c_t$ are respectively \textit{input gate}, \textit{forget gate}, \textit{output gate} and \textit{cell} activation vector at time $t$.

If only hidden state $h_t$ is used for predicting next output at time $t$, it is called unidirectional in that RNN made use of only previous context. However in speech recognition, where whole utterance are transcribed at once, the correct interpretation of the current sound may depend on next few frames as well as previous frames because of the co-articulation.

Bidirectional RNN combines RNN that moves forward through time with another RNN that moves backward through time. So given forward hidden state $\overrightarrow{h}_t$ and backward hidden state $\overleftarrow{h}_t$ at time $t$, the bidirectional hidden state becomes concatenation of them:
\begin{equation}
h_t = [\overrightarrow{h}_t ; \overleftarrow{h}_t]
\end{equation}

\subsection{Optimization}

As nonlinear activation function makes neural network a non-convex function, the neural network is optimized by gradient-based learning algorithms. Stochastic Gradient Descent (SGD) is mostly used optimization algorithms and many others including Adam \cite{kingma2014adam}, which is used for the proposed model, are the variant of SGD.

When the neural network defines a distribution $p(\mathbf{y};\mathbf{x},\theta)$, the cost function of the model is derived from the principle of maximum likelihood and it could be equivalently described as negative log-likelihood or cross-entropy between the training data and the model distribution:

\begin{equation}
J(\theta) = -\mathbb{E}_{\mathbf{x},\mathbf{y}~\bar{p}_{data}} \log p_{model} (\mathbf{y}|\mathbf{x})
\end{equation}

The SGD drives cost function $J(\theta)$ to a very low value iteratively using gradient which is computed by taking the average gradient on a minibatch of $m$ examples randomly chosen from the training set.

\begin{algorithm}
\caption{Stochastic Gradient Descent (SGD)}
\label{alg:sgd}
\begin{algorithmic}
\Require Learning rate $\epsilon$, Initial parameter $\theta$
\While{model is not converged}
\State Sample a minibatch of $m$ examples $\{\mathbf{x}_1, ..., \mathbf{x}_m\}$  with corresponding targets 
\State $\mathbf{y}_i$ from the training set.
\State Compute gradient: $\mathbf{g} \leftarrow \frac{1}{m} \nabla_{\theta} \sum_i L(f(\mathbf{x}_i;\theta),\mathbf{y}_i)$ 
\State Apply update: $\theta \leftarrow \theta - \epsilon \mathbf{g}$
\EndWhile
\end{algorithmic}
\end{algorithm}

\subsection{Regularization}

When the model performs well on training data but not on new inputs it is called overfitting. To prevent overfitting, there are various of regularization algorithms which reduce the generalization error by limiting capacity of models.

The weight decay which is used for the proposed model in the fine-tunning stage, is commonly used regularization technique in the machine learning community. Specifically, the weight decay adds a regularization term $\frac{1}{2} \|\mathbf{w} \|_2^2$ to the cost function $J(\theta)$.

It is also called $L^2$ parameter regularization or ridge regression and has property of driving the weights closer to the origin. So if $\theta$ is $\mathbf{w}$ with assumption of no bias for brevity, the total cost function become:

\begin{equation}
\tilde{J}(\mathbf{w}) = \frac{\alpha}{2} \mathbf{w}^{\top} \mathbf{w} + J(\mathbf{w})
\end{equation}

Where $\alpha$ is hyper-parameter that decides the contribution of regularization term.

\section{Attention-based sequence to sequence model}

Attention-based sequence to sequence model (attention-based seq2seq) is general sequences transducer for mapping input sequences to output sequences with attention mechanism. Here the attention mechanism plays a major role for seq2seq model to utilize whole input sequences effectively when producing output sequences. It is widely used in various domains according to the type of input and output sequences. For example, attention-based seq2seq model translated English to French \cite{Bahdanau2016Neural}, recognize phoneme sequence from speech \cite{chorowski2015attention} and even generated English sentences describing given image \cite{Xu2016Show}, so called image captioning problem.

There are many possible architecture for attention-based seq2seq model and I will describe one of them for speech recognition. Given character sequences $\mathbf{y} = (y_1, y_2, ..., y_T)$ and acoustic feature sequences $\mathbf{x} = (x_1, x_2, ..., x_S)$, the attention-based seq2seq model produce output value one at a time by modeling conditional character probability distribution given whole input sequences and past character sequences $P(y_t|\mathbf{x},y_{<t})$. With chain rule, it models character sequences distribution given acoustic feature sequences as:
\begin{equation}
P(\mathbf{y}|\mathbf{x}) = \prod_t P(y_t|\mathbf{x}, y_{<t})
\end{equation}

In practice, it is divided into two parts; encoder and decoder. The encoder is bidirectional LSTM which consumes input sequences and produces hidden state sequences $\mathbf{h} = (h_1, h_2, ... h_{S'})$ where length of $\mathbf{h}$ is not longer than length of $\mathbf{x}$ ($S' <= S$). The decoder is unidirectional LSTM and it produces output value one at a time until end of sequence label is emitted while utilizing hidden state sequences of encoder with attention mechanism. This procedures can be formulated as follows:

\begin{equation}
\begin{aligned}
\mathbf{h} &= Encoder(\mathbf{x}) \\
P(y_t|\mathbf{x},y_{<t}) &= Decoder(\mathbf{h}, y_{<t}) \\
P(\mathbf{y}|\mathbf{x}) &= \prod_t P(y_t|\mathbf{x}, y_{<t}) \\
\end{aligned}
\end{equation}

The experiments done in this thesis is phoneme recognition with `TIMIT' speech dataset so $\mathbf{x}$ is log mel-scaled filter bank coefficients sequences and $\mathbf{y}$ is its transcribed phoneme sequences. Specific details of how decoder attends to hidden state sequences of encoder are explained in the next section.

\chapter{Model component}

In this section, I will describe each components of the proposed model. Although each algorithms have been developed in different domain, the combined architecture suggested in this thesis showed the significant performance improvement on phoneme recognition.

\section{Luong's attention mechanism}

\begin{figure}[h!]
\centering
\includegraphics[width=\textwidth,height=0.5\textwidth]{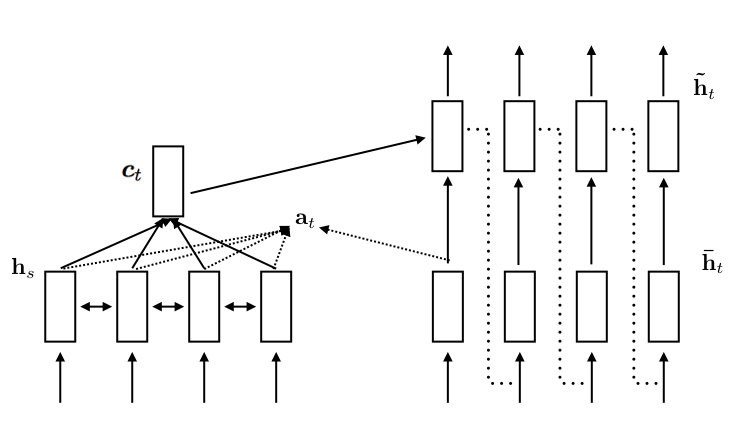}
\caption{The attention-based seq2seq model where encoder is bidirectional and encoder is unidirectional recurrent neural network}
\label{fig:attention_model}
\end{figure}

Luong in \cite{DBLP:journals/corr/LuongPM15} propose simple and effective architecture for attention-based model. The Luong's attention mechanism could be summarized into two parts. First, how to obtain the context vector and output the predictions from it.
Second, the input-feeding approach to inform a model the previous alignments.

The context vector $\mathbf{c}$ is the way the decoder attends to hidden state sequences of encoder. Specifically, at each time step, current hidden state of decoder $\mathbf{\bar{h}}_t$ is compared to each hidden state of encoder $\mathbf{h}_s$ to produce a variable-length alignment vector $\mathbf{a}_t$:

\begin{equation}
\mathbf{a}_t(s) = \frac{exp(\mathbf{\bar{h}}_t^{\top}\mathbf{W}_a \mathbf{h}_s)}{\sum_{s'}exp(\mathbf{\bar{h}}_t^{\top} \mathbf{W}_a \mathbf{h}_{s'})}
\end{equation}

Where $\mathbf{h}$, $\mathbf{\bar{h}}$ denote hidden states at the top of the LSTM layers in the encoder and the decoder each.

At each time step $t$, given this alignment vector $\mathbf{a}_t$ as weights, the context vector $\mathbf{c}_t$ is computed as the weighted average over all the hidden state sequences of encoder $\mathbf{h}$.

Then, context vector $\mathbf{c}_t$ is concatenated with hidden state of decoder $\mathbf{\bar{h}}_t$ to produce attentional hidden state $\tilde{\mathbf{h}}_t$.

\begin{equation}
\mathbf{\tilde{h}}_t = tanh(\mathbf{W}_c [\mathbf{c}_t ; \mathbf{h}_t])
\end{equation}

Finally the attentional vector $\mathbf{\tilde{h}}_t$ is fed into the softmax layer to model conditional character distribution given input data and past character sequences.

\begin{equation}
P(y_t|y_{<t}, \mathbf{x}) = softmax(\mathbf{W}_s \tilde{\mathbf{h}}_t)
\end{equation}

The input-feeding approach is crucial for attention model to successfully align the input sequences and the output sequences. It is implemented by concatenating attentional vector $\mathbf{\tilde{h}}_t$ with the input at the next time step thus previous alignment information is maintained through hidden state of decoder.

The figure \ref{fig:attention_model} depicts a simple sequence to sequence model with Luong's attention mechanism.

\section{Batch normalization}

When training deep neural network, it is good practice to normalize each feature of input data over whole training set since this normalized distribution is relatively easy to train. However as layers become deeper, the distribution of hidden unit's activation changes a lot which make it hard to train deep neural network. The batch normalization \cite{ioffe2015batch} extends this idea further so that each activation of hidden layer is also normalized over the batched dataset. If $\mu_\mathcal{B}$ is batch mean and $\sigma_\mathcal{B}^2$ is batch variance, $i$th activation could be normalized as:
\begin{equation}
\hat{x}_i = \frac{x_i - \mu_\mathcal{B}}{\sqrt{\sigma_\mathcal{B}^2 + \epsilon}}
\end{equation}
Where $\epsilon$ is small value for numerical stability.
The batch normalized activation now have zero mean and unit variance, but sometimes this property is not so powerful thus original representation may be preferred instead. The batch normalization introduce two more parameter $\gamma$ and $\beta$ so that it restore representational power of the network. With this modification, batch normalized $i$th activation is formulated as:
\begin{equation}
y_i = \gamma \hat{x}_i + \beta
\end{equation}

When it is used in convolutional neural network, to preserve convolutional property, the normalization occurs over all location of feature map where the activation is contained as well as over the batched data.

Since the proposed model in this thesis have deep encoder with stacked convolutional layer, to make it trainable, it was crucial to apply batch normalization at every neural network layer in the encoder part.

\section{Dropout}

Dropout \cite{Srivastava:2014:DSW:2627435.2670313} is powerful regularization technique which has effect of bagged ensemble models. The bagged ensemble is well known regularization technique where several models trained on separately are combined to output the predictions. However, training neural network is costly in terms of time and memory so bagged ensemble of neural networks is impractical. 

\begin{figure}[h]
\centering
\includegraphics[width=0.5\textwidth,height=0.5\textwidth]{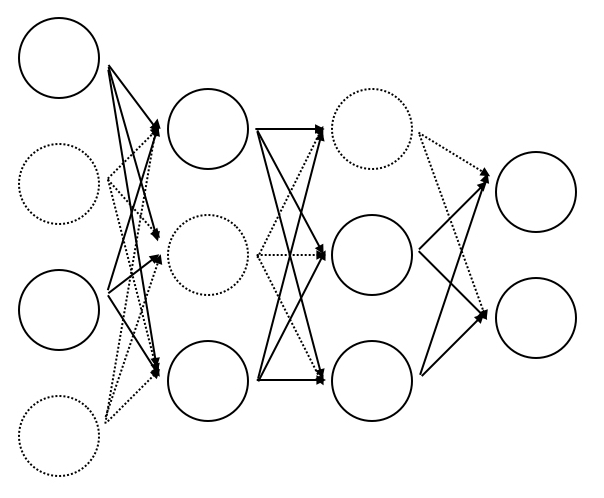}
\caption{The one of the subnetwork with dropped units.}
\label{fig:dropout}
\end{figure}

In dropout, instead of training several model separately, exponentially many sub networks which are formed by removing nonoutput units from an underlying base network are trained. As shown in figure \ref{fig:dropout}, different sub network is trained per each train step.

In inference time, to approximate predictions of ensemble models, the model of all units is used with all units multiplied by the probability of including units. It is called weight scaling inference rule and motivation of it is to approximate the expected value of the output from that unit.

To prevent overfitting in early training stage, I used dropout after every neural network layers which includes convolutional, fully connected, LSTM and attentional layer.

\section{Residual network}

The deep neural network have been known to be better than shallow one as it can obtain more useful representations in the hierarchy of concepts. However more layers occasionally degrade the performance in training dataset which is somewhat counter-intuitive since degradation of training error could be avoided if residual layer just learn identity mapping. This indicates that some architectures are easier to optimize than others.

The residual network \cite{he2016deep} suggest that few stacked layer should learn residual mapping instead of directly fitting a desired underlying mapping. For example, if underlying mapping is $\mathcal{H}(\mathbf{x})$, stacked layers would learn residual function $\mathcal{F}(\mathbf{x}) = \mathcal{H}(\mathbf{x}) - \mathbf{x}$ which is expected to be easier to learn.

A few stacked convolutional layer with residual mapping is widely used neural network architecture in computer vision community. So it is also adapted for the proposed model with some modification.

\chapter{Experiment}

\section{Proposed model architecture}

\subsection{Design principle}
The proposed model is differ from conventional attention-based seq2seq model in that it has deep stacked neural network in encoder part. This design decision was made with two principles.

First, the speech has structured property over frequency and time axis like image data. This property may be effectively exploited by two-dimensional convolutional layer so applying it before LSTM layer is expected to give more useful representation for next LSMT layer compared to directly consuming log mel-scaled filter bank feature vectors.

Second, I assumed that deep encoder would be better than shallow one. Since deep convolutional layers are hard to train, it was crucial to apply batch normalization and residual network for deep encoder to be converged.

\subsection{Building blocks}
The proposed model has deep stacked layers in encoder part and it can be grouped into three logical blocks.

\begin{figure}[h!]
\centering
\includegraphics[width=0.5\textwidth,height=0.5\textwidth]{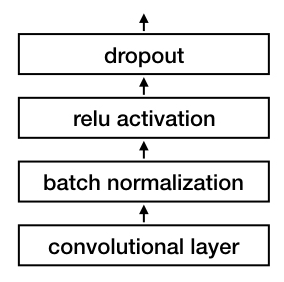}
\caption{The conv block}
\label{fig:conv_block}
\end{figure}

\begin{figure}[h!]
\centering
\includegraphics[width=0.5\textwidth,height=0.6\textwidth]{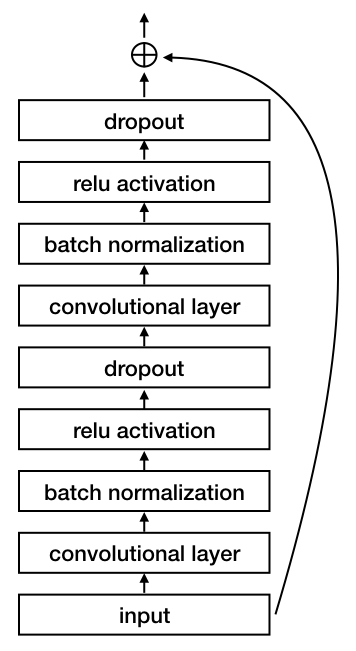}
\caption{Residual block: input is the output of previous layer}
\label{fig:residual_block}
\end{figure}

\begin{figure}[h!]
\centering
\includegraphics[width=0.5\textwidth,height=0.5\textwidth]{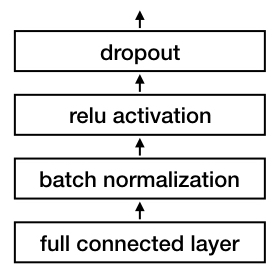}
\caption{The dense block}
\label{fig:dense_block}
\end{figure}

The conv block shown in figure \ref{fig:conv_block}, is first convolutional layer which consumes input data. It reduces time resolution with stride of $1 \times 3$ which means there are no reduction in frequency axis whereas the length of input sequences reduced in a factor of three. Without time reduction, GPU ran out of memory before stacking deep encoder. Moreover it speed up training time by making LSTM layer in upper layers process the reduced sequence data.

The residual block in figure \ref{fig:residual_block}, consists of two convolutional neural network with residual mapping. It is similar architecture to conventional residual network except two differences. Those are inclusion of Dropout and the order of residual addition operation. For intensive regularization, Dropout is applied after each relu activation. And the input is simply added to the output of residual network instead of adding it before relu activation.

Figure \ref{fig:dense_block} depicts dense block which is just feed forward neural network with Bath normalization and Dropout. It is stacked after residual block and before LSTM layer in encoder. Since the flattend output of residual block has too many units compared to the LSTM units in the proposed model, dense block is inserted between them for making compact representation. This method is common in image recognition where stacked convolutional layers are followed by fully connected layers.

\subsection{Details of final architecture}

\begin{figure}[h!]
\centering
\includegraphics[width=0.5\textwidth,height=0.75\textwidth]{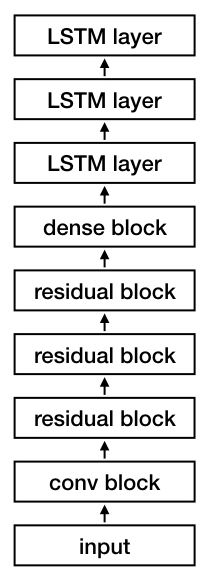}
\caption{The encoder: Dropout which is applied at each output of LSTM layers is omitted for brevity.}
\label{fig:encoder}
\end{figure}

\begin{figure}[h!]
\centering
\includegraphics[width=0.5\textwidth,height=0.5\textwidth]{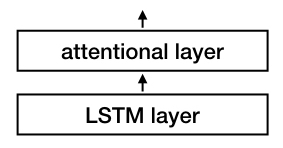}
\caption{The decoder: Dropout which is applied to output of LSTM and attentional layer is omitted for brevity.}
\label{fig:decoder}
\end{figure}

The proposed model has deep architecture in encoder part as shown in figure \ref{fig:encoder}. The deep encoder has one conv block, three residual blocks and one dense block followed by three LSTM layers. All convolutional layers have kernel size of $3 \times 3$ and stride of $1 \times 1$ except the one at conv block where stride is $1 \times 3$ for time reduction. The convolutional layer in residual block has 64 feature maps whereas 128 feature maps are used in conv block. The dense block has 1024 units and bidirectional LSTM layers have 256 unit in each direction.

Figure \ref{fig:decoder} depicts the decoder where unidirectional LSTM layer and attentional layer both have 256 units.

\section{Data description}

I evaluated the proposed model on the `TIMIT'. It is commonly used benchmark dataset for speech recognition and consists of recorded speech and orthographic transcription. Following the data split procedure \cite{halberstadt1999heterogeneous}, the proposed model was trained on the standard 462-speaker training set with all SA records removed. The 50-speaker development set was used for early stopping. Final evaluation was performed on the core test set including 192 sentences. 

\section{Training details}

The input features were 40 log mel-scaled filter bank (plus energy term) coefficients with deltas and deltas-deltas, which results in 123 dimensional features. Each dimension was normalized to have zero mean and unit variance over the training set. It was then reshaped to $(41 \times timestep \times 3)$ for convolutional layer where second and third channels were delta and delta-delta. The phoneme consumed by the decoder at each time step was one-hot encoded whereas zero-valued vector was used for start-of-sequences label.

Training and decoding were done on full 61 phone labels plus end-of-sequences label which was appended to each target sequences whereas scoring was done on 39 phoneme set. Decoding is performed with a simple left-to-right beam search algorithm with beam width 10. Specifically at each time step, each partial hypothesis was expanded with every possible output value and only 10 most likely hypotheses was maintained until the end-of-sequences label is encountered. Among 10 possible hypotheses, the one with most high probability was chosen as final transcription.

To optimize the proposed model, Adam \cite{kingma2014adam} algorithm with learning rate $10^{-3}$, default parameter and batch size 32 was used. The dropout rate of 0.5, gradient norm clipping to 1 were applied for training. After it converged I fine-tunned by decaying learning rate to $10^{-4}$ and adding weight decay $10^{-5}$ regularization. All weight matrices were initialized according to glorot uniform \cite{pmlr-v9-glorot10a} except LSTM weight matrices which were initialized from uniform distribution $\mathcal{U}(-0.1, 0.1)$.

\section{Performance evaluation metric}
When model's output is sequences rather than discrete class label, typical classification accuracy metrics could not be used. Instead $edit$ or $Levenshtein$ distance between sequences is used for performance evaluation metric. For example, if $\mathbf{y}_{true}$ is a true word sequences and $\mathbf{y}_{pred}$ is predictions of model the error metric which is called word error rate (WER) in this case is calculated as:

\begin{equation}
WER = \frac{1}{Z} \sum_{(\mathbf{y}_{true}, \mathbf{y}_{pred}) \in \mathbf{S}} ED(\mathbf{y}_{true}, \mathbf{y}_{pred})
\end{equation}

Where $Z$ is total length of $\mathbf{y}_{true}$ in test set and $\mathbf{S}$ is test set including all $\mathbf{y}_{true}$, $\mathbf{y}_{pred}$ pairs.

In the case of phoneme recognition like `TIMIT', each sequences consist of phoneme so it is called phoneme error rate (PER).

\section{Results}

The proposed model, as far as I know, has achieved state-of-the-art results of phoneme error rate. The table \ref{table:results} compares previous results on `TIMIT' dataset.

\begin{table}[h]
	\centering
	\caption{Phoneme Error Rate (PER) on TIMIT}
    \label{table:results}
	\begin{tabular}{lc}
		\toprule
        Model & PER \\
        \midrule
        RNN transducer \cite{Graves2013Speech} & 17.7\% \\
        Attention-based seq2seq \cite{chorowski2015attention} & 17.6\% \\
        HMM over time and frequency convolutional net \cite{6853584} & 16.7\% \\
        Convolutional attention-based seq2seq (proposed model) & \textbf{15.8\%}
	\end{tabular}
\end{table}

\chapter{Conclusions}

In this thesis, I introduced the automatic speech recognition (ASR) and basic of neural network and attention-based sequence to sequence model. Then I described each components of the model including Luong's attention mechanism, Dropout, Batch normalization and Residual network. Finally, I proposed the convolutional attention-based sequence to sequence model for end-to-end ASR.

The proposed model was based on seq2seq model with Luong's attention mechanism and it built deep encoder with several stacked convolutional neural network. This deep architecture could have been trained successfully with Batch normalization, Residual network and powerful regularization of Dropout. 

I experimentally proved the superiority of the convolutional attention-based seq2seq neural network by showing state-of-the-art results on phoneme recognition.

In the future, I hope to build deeper network based on the proposed model with more powerful computing resources. So the very deep convolutional attention-based seq2seq model trained on large vocabulary datasets might be interesting study. Another interesting direction would be to build the proposed model with pure convolutional neural networks by removing the recurrent neural networks completely in the encoder. The removal of RNN could speed up training and inference time of the model by paralleling the convolution operation on graphic processing unit (GPU) efficiently.

\nocite{Rabiner1989Tutorial}
\nocite{Goodfellow-et-al-2016}
\nocite{tensorflow2015-whitepaper}
\bibliographystyle{unsrt}
\bibliography{imdanboy}
\end{document}